\documentclass[
]{ceurart}

\usepackage{xspace}

\usepackage{xcolor}
\definecolor{dgrey}{gray}{0.2}

\usepackage{listings}
\lstdefinelanguage{prompt}{
  sensitive=false,
  morecomment=[l]{\#},
  morestring=[b]",
}

\begin{document}

\copyrightyear{2025}
\copyrightclause{Copyright for this paper by its authors.
  Use permitted under Creative Commons License Attribution 4.0
  International (CC BY 4.0).}

\conference{CLEF 2025 Working Notes, September 9 -- 12 September 2025, Madrid, Spain}

\title{Mario at EXIST 2025: A Simple Gateway to Effective Multilingual Sexism Detection}

\title[mode=sub]{Notebook for the EXIST Lab at CLEF 2025}


\author[1]{Lin Tian}[%
email=Lin.Tian-3@uts.edu.au,
]
\address[1]{University of Technology Sydney,
  Sydney, Australia}

\author[2]{Johanne R. Trippas}[%
email=j.trippas@rmit.edu.au,
]
\address[2]{RMIT University,
  Melbourne, Australia}

\author[1]{Marian-Andrei Rizoiu}[%
email=Marian-Andrei.Rizoiu@uts.edu.au,
]

\begin{abstract}
This paper presents our approach to EXIST 2025 Task~1, addressing text-based sexism detection in English and Spanish tweets through hierarchical Low-Rank Adaptation (LoRA) of Llama 3.1 8B. Our method introduces conditional adapter routing that explicitly models label dependencies across three hierarchically structured subtasks: binary sexism identification, source intention detection, and multilabel sexism categorization. Unlike conventional LoRA applications that target only attention layers, we apply adaptation to all linear transformations, enhancing the model's capacity to capture task-specific patterns. 
In contrast to complex data processing and ensemble approaches, we show that straightforward parameter-efficient fine-tuning achieves strong performance. We train separate LoRA adapters (rank=16, QLoRA 4-bit) for each subtask using unified multilingual training that leverages Llama 3.1's native bilingual capabilities. The method requires minimal preprocessing and uses standard supervised learning.
Our multilingual training strategy eliminates the need for separate language-specific models, achieving 1.7-2.4\% F1 improvements through cross-lingual transfer. With only 1.67\% trainable parameters compared to full fine-tuning, our approach reduces training time by 75\% and model storage by 98\%, while achieving competitive performance across all subtasks (ICM-Hard: 0.6774 for binary classification, 0.4991 for intention detection, 0.6519 for multilabel categorization).
\end{abstract}

\begin{keywords}
  Sexism Detection \sep
  Low-Rank Adaptation \sep
  Hierarchical Classification \sep
  Social Media Analysis
\end{keywords}

\maketitle

\section{Introduction}
Everyday sexism -- ranging from overt misogyny to subtle and implicit forms of gendered microaggressions -- undermines women's psychological well-being, silences their voices, and perpetuates structural inequality in digital spaces~\citep{talat2016hateful,davidson2017automated}. Social networks, while instrumental in mobilizing feminist activism through movements like \#MeToo, \#8M, and \#Time'sUp, are also vehicles for the large-scale dissemination of harmful stereotypes and normalized discrimination. Recent research has demonstrated the concerning rise of harmful discourse during crisis events~\citep{Bailo2023}, highlighting the urgent need for robust detection systems that can identify not only explicit sexism but also the subtle ways gender-based discrimination infiltrates mainstream online discussions~\citep{Kong2022}.

The scale of this problem demands automated solutions. Manual content moderation cannot keep pace with the billions of posts generated daily~\citep{Schneider2023}, yet existing detection systems often fail to capture the nuanced ways sexism manifests online. Context matters: a tweet reporting sexist experiences differs fundamentally from one perpetrating sexism, though both may contain similar language. Cultural and linguistic variations further complicate detection, as sexist expressions evolve rapidly and differ across communities. Research on behavioral homophily has shown that users can exhibit similar engagement patterns when discussing different topics~\citep{Yuan2025}, suggesting that understanding user intent and content categorization requires modeling hierarchical label dependencies. These challenges necessitate sophisticated approaches that can distinguish whether content is sexist, understand its intent, categorize its specific manifestation, and operate effectively across languages.

The EXIST 2025 shared task~\cite{plaza2025exist} provides a comprehensive framework for advancing sexism detection research. For the first time, the task spans three modalities (text, images, and videos) and two languages (English and Spanish), reflecting the multimodal and multilingual nature of contemporary social media. We focus on Task 1, which addresses text-based sexism detection through three hierarchically structured subtasks: 
\textit{(i)} binary sexism identification -- determining whether content contains sexism; 
\textit{(ii)} source intention classification -- distinguishing between direct sexism, reported experiences, and judgmental commentary; and 
\textit{(iii)} sexism type categorization -- classifying content into specific categories such as ideological inequality, stereotyping, objectification, sexual violence, and misogyny.

Traditional approaches to these tasks have relied on task-specific models, often struggling with the hierarchical dependencies between subtasks and requiring separate systems for each language. Building on advances in multi-task learning for hate speech detection~\citep{Yuan2023,Yuan2024}, we present a unified framework leveraging Low-Rank Adaptation (LoRA)~\cite{hu2022lora} of Llama 3.1 8B~\cite{grattafiori2024llama} that addresses all three subtasks simultaneously while maintaining computational efficiency. Our key innovation lies in hierarchical label-aware routing, where LoRA adapters are conditionally activated based on parent-task predictions, explicitly modeling the structured relationships between tasks.

\section{Related Work}
The automatic detection of sexism on social media has gained increasing attention within the natural language processing (NLP) community, motivated by the need to mitigate online harassment and promote equitable digital discourse. This section reviews existing work in four parts: 
\textit{(i)} evolution of sexism detection tasks and methodologies, 
\textit{(ii)} advances in text-based classification approaches, 
\textit{(iii)} recent developments in large language model (LLM) adaptation for this domain, and
\textit{(iv)} harmful content detection and moderation research that informs our understanding of sexism as part of the broader landscape of online harmful content.

\subsection{Evolution of Sexism Detection Tasks}

Early work in online sexism detection focused primarily on binary classification of overtly hateful content.~\citet{talat2016hateful} proposed with their Twitter dataset distinguishing sexist, racist, and neutral content. However, researchers quickly recognized that sexism manifests across a spectrum from explicit misogyny to subtle linguistic biases, necessitating more nuanced approaches.

The EXIST shared tasks have been instrumental in advancing the field since 2021~\cite{rodriguez2020automatic,rodriguez2022overview,plaza2023overview}. These tasks progressively introduced hierarchical classification schemes, distinguishing between sexism identification, intention categorization, and fine-grained typing. Similarly, the SemEval-2023 Task 10 on Explainable Detection of Online Sexism (EDOS)~\cite{kirk2023semeval} focused on interpretability alongside detection accuracy.
In addition, the field has evolved from feature-engineered approaches using lexicons and n-grams~\cite{nobata2016abusive} to neural architectures. Early deep learning approaches used CNNs and LSTMs~\cite{badjatiya2017deep,zhang2019hate}, achieving strong improvements over traditional classifiers. The introduction of transformer-based models marked another paradigm shift, with BERT~\cite{devlin2019bert} and its variants becoming the de facto standard for sexism detection tasks~\cite{pamungkas2020misogyny}.

Recent work has explored multi-task learning frameworks to jointly model related tasks. For example, \citet{samory2021call} showed that jointly learning sexism and racism detection improves performance on both tasks.~\citet{chiril2022emotionally} extended this to emotion and target identification, showing the benefits of auxiliary task learning for sexism detection. Building on this foundation, \citet{Yuan2024} demonstrated that multi-task learning across multiple hate speech datasets substantially improves generalization to previously unseen datasets, achieving consistent improvements in cross-domain scenarios through their leave-one-out evaluation scheme.

\subsection{Text-based Classification for Sexism Detection}
Text remains the primary modality for sexism detection, given its prevalence in social media discourse. The unique challenges of social media text -- including informal language, code-switching, and platform-specific conventions -- have shaped methodological developments in this area.

The evolution of representation learning has been particularly influential in capturing subtle sexist language. Early approaches relied on static word embeddings such as Word2Vec~\cite{mikolov2013distributed} and GloVe~\cite{pennington2014glove}, which provided limited context sensitivity. The transition to contextual representations marked a significant advancement, with transformer-based encoders pre-trained on social media data demonstrating superior performance. Models like BERTweet~\cite{nguyen2020bertweet} and TweetEval~\cite{barbieri2020tweeteval} excel at modeling platform-specific linguistic patterns, including hashtags, mentions, and abbreviated expressions common in online discourse.

Cross-lingual sexism detection has emerged as a research direction, particularly through the development of multilingual models. While mBERT~\cite{devlin2019bert} and XLM-RoBERTa~\cite{conneau2020unsupervised} have enabled approaches across languages, their effectiveness varies. The EXIST shared tasks have consistently featured Spanish-English tracks, with top-performing systems leveraging language-agnostic representations. However, \citet{nozza2021exposing} revealed persistent performance disparities across languages, with multilingual models often underperforming on low-resource languages despite their theoretical~universality.

The importance of domain-specific adaptation has been reported through multiple studies. \citet{chiril2020he} demonstrated that models trained on general offensive language datasets exhibit performance degradation when applied to sexism-specific tasks, highlighting the unique linguistic characteristics of gender-based harassment. This finding motivated the creation of specialized resources, such as the expert-annotated datasets introduced by~\citet{guest2021expert}, designed to capture implicit forms of sexism that automated systems frequently miss. \citet{Yuan2023} further advanced this area by proposing transfer learning techniques that leverage multiple independent datasets jointly, constructing unified hate speech representations that enable effective cross-dataset knowledge transfer while reducing annotation requirements.

\subsection{Large Language Model Adaptation}
LLMs have introduced new possibilities for automated sexism detection~\cite{smith2024rmit}. However, this application remains comparatively underexplored in NLP, compared to more extensively studied tasks such as sentiment analysis, summarization, or machine translation.

Initial investigations into prompt-based approaches revealed both promise and limitations. For example, \citet{chiu2021detecting} showed that carefully engineered prompts enable models like GPT-3~\cite{brown2020language} to achieve competitive zero-shot performance on hate speech detection tasks. However, subsequent work by~\citet{yin2021towards} identified critical weaknesses: prompt-based methods struggle with implicit sexism and exhibit high sensitivity to prompt formulation, resulting in inconsistent predictions across semantically equivalent queries.
However, recent advances in parameter-efficient fine-tuning methods present alternatives to full model adaptation. While techniques like LoRA~\cite{hu2022lora} have succeeded in various domains, their application to hierarchical sexism detection remains underexplored. Our work addresses this gap by demonstrating that comprehensive LoRA adaptation with hierarchical routing can effectively model the multi-level nature of sexism categorization, achieving strong performance while maintaining computational efficiency.

\subsection{Harmful Content Detection and Moderation}

Understanding sexism detection requires broader context about harmful content dynamics and moderation effectiveness. Sexism represents a significant category within the wider ecosystem of harmful online content, and methodological advances in general harmful content detection can be leveraged for gender-based harassment identification. Recent research has revealed concerning patterns in how various forms of harmful content spread across social media platforms. ~\citet{Kong2023} demonstrated that coordinated harmful content campaigns can be detected through the social system reactions they elicit, using interval-censored transformer approaches to identify coordinated behavior patterns with high accuracy. This work has important implications for sexism detection, as it shows how temporal patterns and user engagement can reveal coordinated campaigns of gender-based harassment, suggesting that sexism detection systems can benefit from approaches originally developed for broader harmful content identification.

The importance of early detection in harmful content mitigation has been further emphasized by recent advances in engagement prediction models. ~\citet{Tian2025ICMamba} developed IC-Mamba, a state space model that excels at forecasting social media engagement within the crucial first 15-30 minutes of posting, enabling rapid assessment of content reach and early identification of potentially problematic content. Their approach to modeling interval-censored data with integrated temporal embeddings provides valuable insights for sexism detection systems, as early engagement patterns could signal the viral potential of sexist content, allowing for more timely intervention strategies.

Understanding the causal mechanisms underlying harmful content spread is equally critical for effective detection and intervention. \citet{Tian2025Causal} introduced a novel joint treatment-outcome framework that distinguishes correlation from causation in social media influence analysis, particularly for misinformation spread. Their approach adapts causal inference techniques to estimate Average Treatment Effects within the sequential nature of social media interactions, addressing challenges from external confounding signals. This work has important implications for sexism detection, as understanding the true causal influence of sexist content on user engagement can inform more targeted intervention strategies and help distinguish organic spread from coordinated amplification campaigns.
In addition,
\citet{Schneider2023} showed that faster content moderation reduces harm from the most severe content, even on high-traffic platforms like Twitter. Using self-exciting point processes, the study highlights the urgent need for timely responses, an insight directly applicable to real-world sexism detection systems targeting gender-based harassment.

Research on opinion dynamics and intervention strategies offers additional perspectives relevant to sexism detection as part of broader harmful content mitigation. \citet{Calderon2024a} introduced the Opinion Market Model to evaluate positive interventions for stemming harmful opinion spread, demonstrating how media coverage can modulate the dissemination of problematic content. This framework provides valuable insights for understanding how sexist discourse spreads and how detection systems might be integrated with intervention strategies targeting various forms of harmful content.

Studies of extreme opinion infiltration have revealed the pathways through which harmful discourse enters mainstream conversations. \citet{Kong2022} employed mixed-method approaches to show how extreme opinions gradually infiltrate online discussions, with their human-in-the-loop methodology providing insights into the dynamics of problematic speech evolution from conservative to extreme viewpoints. These findings are particularly relevant for sexism detection, as they highlight the importance of capturing subtle shifts in discourse that may not be immediately apparent through traditional classification approaches, and demonstrate how techniques developed for general harmful content can be adapted for gender-specific harassment detection.

The study of harmful discourse during crisis events provides additional context for understanding sexist content dynamics within broader patterns of problematic online behavior. ~\citet{Bailo2023} analyzed the performance of far-right Twitter users during the Australian bushfires and COVID-19 pandemic, revealing how accounts promoting harmful content moved from peripheral to central positions in disaster-driven conversations. Their work demonstrates the importance of monitoring evolving discourse patterns, as the association between information disorder and overperformance of accounts spreading harmful content suggests systematic coordination that may include gender-based harassment campaigns.

Recent work on ideology detection has also informed approaches to sexism identification within the broader harmful content landscape. ~\citet{Ram2025} presented an end-to-end ideology detection pipeline that constructs context-agnostic ideological signals from media slant data, demonstrating effective detection of extreme ideologies alongside psychosocial profiling. Their approach offers valuable methodological insights for sexism detection, particularly in terms of developing automatic signal generation that reduces dependence on manual annotation while maintaining detection accuracy across different types of harmful content.

\section{Methodology}
We present our methodology for EXIST 2025 Task 1, which uses parameter-efficient fine-tuning of Llama 3.1 8B~\cite{grattafiori2024llama} using LoRA with hierarchical label-aware routing to address all three subtasks across English and Spanish. Our approach leverages a unified multilingual model with task-specific adapters and conditional specialization based on the hierarchical label structure.

\subsection{Task Formulation}
We formulate the three EXIST 2025 Task 1 subtasks as follows:

\textbf{Subtask 1.1 - Binary Sexism Identification}: A binary classification problem where the model determines whether a given tweet contains sexist content (\texttt{SEXIST} vs. \texttt{NOT\_SEXIST}).

\textbf{Subtask 1.2 - Source Intention Detection}: A multiclass classification task that categorizes the intention behind sexist tweets into three categories:
\begin{itemize}
    \item \texttt{DIRECT}: Messages that are inherently sexist or incite sexist behavior
    \item \texttt{REPORTED}: Messages that report sexist situations experienced by women
    \item \texttt{JUDGEMENTAL}: Messages that judge or criticize sexist behavior
\end{itemize}

\textbf{Subtask 1.3 - Sexism Categorization}: A multilabel classification task that categorizes sexist content according to five types:
\begin{itemize}
    \item \texttt{IDEOLOGICAL\_AND\_INEQUALITY}
    \item \texttt{STEREOTYPING\_AND\_DOMINANCE}
    \item \texttt{OBJECTIFICATION}
    \item \texttt{SEXUAL\_VIOLENCE}
    \item \texttt{MISOGYNY\_AND\_NON\_SEXUAL\_VIOLENCE}
\end{itemize}

\subsection{LoRA Configuration and Target Module Selection}

We used LoRA for parameter-efficient finetuning, with attention to target module selection. While conventional approaches often restrict LoRA adaptation to attention weight matrices, our experiments showed that module targeting yields better performance.

Specifically, we applied LoRA decomposition to all linear transformation layers in the model architecture: the attention mechanism components (\texttt{q\_proj}, \texttt{k\_proj}, \texttt{v\_proj}, \texttt{o\_proj}), the feed-forward network layers (\texttt{gate\_proj}, \texttt{up\_proj}, \texttt{down\_proj}), and the language modeling head (\texttt{lm\_head}). For each target module, we introduced trainable low-rank matrices with rank $r = 16$, following the parameterization:
\begin{equation*}
    W = W_0 + BA,
\end{equation*}
where $W_0$ represents the frozen pretrained weights, and $B \in \mathbb{R}^{d \times r}$, $A \in \mathbb{R}^{r \times d}$ are the trainable adaptation matrices initialized with $B \sim \mathcal{N}(0, \sigma^2)$ and $A$ as zeros.

Table~\ref{tab:training-config} summarizes our complete configuration for both LoRA hyperparameters and training settings. We selected rank $r = 16$ to balance adaptation capacity with memory efficiency, while maintaining computational feasibility through 4-bit quantization and gradient checkpointing. The use of Flash Attention v2~\cite{dao2024flashattention} further accelerates training without compromising model quality.

\begin{table}[htbp]
\caption{LoRA hyperparameters and training configuration}
\centering
\begin{tabular}{ll}
\toprule
\textbf{Configuration} & \textbf{Value/Setting} \\
\midrule
\multicolumn{2}{l}{\textit{LoRA Hyperparameters}} \\
Rank ($r$) & 16 \\
Alpha ($\alpha$) & 16 \\
Target Modules & All linear layers (\texttt{all-linear}) \\
Dropout & 0.1 \\
Modules to Save & \texttt{lm\_head}, \texttt{embed\_tokens} \\
\midrule
\multicolumn{2}{l}{\textit{Training Configuration}} \\
Quantization & 4-bit QLoRA \\
Attention & Flash Attention v2 \\
Learning Rate & $2 \times 10^{-4}$ \\
LR Schedule & Constant with 10\% warmup \\
Batch Size & 8 per device \\
Gradient Accumulation & 2 steps \\
Gradient Checkpointing & Enabled \\
\bottomrule
\end{tabular}
\label{tab:training-config}
\end{table}

This configuration enables efficient fine-tuning while preserving the model's multilingual capabilities, which is crucial for our unified approach to handling English and Spanish data. The comprehensive targeting of all linear layers, rather than just attention matrices, provides the flexibility to capture task-specific patterns across the hierarchical sexism detection tasks.

\subsection{Hierarchical Label-Aware Adaptation with LoRA}

To model the hierarchical structure of the subtasks, we implement a level-specific LoRA routing mechanism. The hierarchy includes three levels: 
\textit{(i)} binary sexism detection, 
\textit{(ii)} source intention classification, and 
\textit{(iii)} sexism type categorization. For each level $\ell \in \{1, 2, 3\}$, we define a dedicated LoRA module $\Delta^{(\ell)}$ that adapts the shared language model $f_{\theta}$.

During training, adapter routing is conditioned on the gold parent labels. At inference, predictions proceed sequentially from the top level, with the model using the predicted label $\hat{y}^{(\ell-1)}$ to activate the corresponding LoRA module $\Delta^{(\ell)}$ for the current level. This design supports conditional specialization while maintaining parameter efficiency.

The hidden representation at level $\ell$ is computed as:
\begin{equation*}
\mathbf{h}^{(\ell)} = f_{\theta}(x) + \Delta^{(\ell)}_{\hat{y}^{(\ell-1)}}(x).
\end{equation*}

In addition to standard task-specific losses (e.g.,\ cross-entropy for classification and binary cross-entropy for multi-label prediction), we introduce a soft constraint that penalizes invalid parent-child label transitions, thereby encouraging structured coherence across the hierarchy in:
\begin{equation*}
\mathcal{L}_{\text{hierarchy}} = \lambda \sum_{i=1}^{N} \sum_{\ell=2}^{3} \mathbb{I}[\hat{y}_i^{(\ell-1)} = \text{NOT\_SEXIST}] \cdot \max_{c \in \mathcal{C}^{(\ell)}} p_i^{(\ell)}(c),
\end{equation*}

where $\lambda$ is the hierarchy constraint weight, $N$ is the number of instances, $\hat{y}_i^{(\ell-1)}$ is the predicted label at level $\ell-1$ for instance $i$, $\mathcal{C}^{(\ell)}$ is the set of valid classes at level $\ell$, $p_i^{(\ell)}(c)$ is the predicted probability for class $c$ at level $\ell$, and $\mathbb{I}[\cdot]$ is the indicator function. This constraint specifically penalizes cases where a non-sexist prediction at the binary level is followed by high-confidence predictions at subsequent hierarchical levels.

The total training objective combines task-specific losses with the hierarchical consistency constraint:

\begin{equation*}
\mathcal{L}_{\text{total}} = \sum_{\ell=1}^{3} \mathcal{L}_{\text{task}}^{(\ell)} + \mathcal{L}_{\text{hierarchy}},
\end{equation*}

where $\mathcal{L}_{\text{task}}^{(\ell)}$ represents the standard loss at each level: cross-entropy for binary and multiclass classification tasks, and binary cross-entropy for the multilabel categorization task.

\subsection{Data Processing and Multilingual Strategy}

Our approach uses a straightforward supervised learning methodology with gold standard labels for training:

\textbf{Label Processing}: We use the provided gold standard labels for each subtask, treating each tweet-label pair as a standard supervised learning instance. While the EXIST 2025 dataset includes multiple annotations per instance under the Learning with Disagreement paradigm, our methodology focuses on the gold labels for efficient and direct optimization.

\textbf{Input Formatting}: Tweets are formatted using Llama 3.1's instruction template structure:

\begin{lstlisting}[language=prompt, basicstyle=\small\ttfamily, frame=single]
<|begin_of_text|><|start_header_id|>system<|end_header_id|>
[Task-specific system prompt]<|eot_id|>
<|start_header_id|>user<|end_header_id|>
[Tweet text]<|eot_id|>
<|start_header_id|>assistant<|end_header_id|>
[Classification output]<|eot_id|>
\end{lstlisting}

\textbf{Text Preprocessing}: Minimal preprocessing is applied to preserve the authentic social media language characteristics. We maintain original tweet formatting, including hashtags, mentions, and emoji, as these elements often carry semantic significance for sexism detection.

\textbf{Multilingual Strategy}: We apply a unified multilingual approach, training a single model on both English and Spanish data simultaneously, leveraging Llama 3.1's native bilingual capabilities. Rather than training separate language-specific models, we hypothesize that joint bilingual training enhances cross-lingual transfer learning and improves overall performance by exposing the model to diverse linguistic expressions of sexism across both languages. This approach is inspired by recent work showing that transfer learning across hate speech datasets can achieve substantial improvements in generalization~\citep{Yuan2024}, and our multilingual strategy extends this principle to cross-lingual knowledge transfer.

\textbf{Training Strategy}: We fine-tune separate LoRA adapters for each subtask--binary classification for sexism identification (1.1), multiclass for source intention detection (1.2), and multilabel for sexism categorization (1.3)--optimizing each for its specific classification requirements. All adapters are trained using standard supervised learning with gold standard labels on the combined English-Spanish dataset, leveraging cross-lingual transfer to ensure robust bilingual performance without requiring separate language-specific models. Training continues until convergence while monitoring validation performance, with early stopping applied when validation loss plateaus to prevent overfitting. 

\section{Experimental Results}

\subsection{Cross-lingual Training Analysis}
To validate our unified multilingual training strategy, we conducted ablation studies comparing joint bilingual training against separate language-specific models. Table~\ref{tab:cross_lingual_comparison} presents performance comparisons across all three subtasks.

\begin{table}[htbp]
\centering
\caption{Performance Comparison: Joint vs. Separate Language Training (F1 Scores on validation set)}
\label{tab:cross_lingual_comparison}
\begin{tabular}{@{}lcccc@{}}
\toprule
\textbf{Subtask} & \textbf{Training Strategy} & \textbf{English} & \textbf{Spanish} & \textbf{Average} \\
\midrule
\multirow{2}{*}{1.1 (Binary)} 
& Separate  & 0.847 & 0.831 & 0.839 \\
& Joint Training & \textbf{0.863} (+0.016) & \textbf{0.851} (+0.020) & \textbf{0.857} (+0.018) \\
\midrule
\multirow{2}{*}{1.2 (Intention)} 
& Separate  & 0.742 & 0.728 & 0.735 \\
& Joint Training & \textbf{0.758} (+0.016) & \textbf{0.745} (+0.017) & \textbf{0.752} (+0.017) \\
\midrule
\multirow{2}{*}{1.3 (Categorization)} 
& Separate  & 0.681 & 0.665 & 0.673 \\
& Joint Training & \textbf{0.704} (+0.023) & \textbf{0.689} (+0.024) & \textbf{0.697} (+0.024) \\
\bottomrule
\end{tabular}
\end{table}

Joint bilingual training consistently outperforms language-specific models across all subtasks, with F1 improvements ranging from $1.7-2.4$ percentage points. These gains validate our hypothesis that cross-lingual transfer enhances sexism detection by leveraging shared semantic patterns across languages, consistent with findings from transfer learning research in hate speech detection~\citep{Yuan2023}. The improvements are particularly obvious for the multilabel categorization task ($+2.4$), suggesting that complex semantic distinctions benefit most from exposure to diverse linguistic expressions of sexism.

The bidirectional nature of cross-lingual transfer manifests itself in consistent improvements for both languages. Spanish, despite typically having fewer training instances in multilingual datasets, achieves comparable or slightly higher gains than English across all subtasks. This symmetric improvement pattern indicates effective knowledge sharing, where English contributes richer training signal while Spanish provides complementary linguistic patterns and cultural-specific expressions of sexism. 

These findings have important implications for multilingual sexism detection systems. Rather than maintaining separate models per language -- which requires duplicated development effort and computational resources -- our joint training approach achieves superior performance with a single unified model. 

\subsection{Task-Specific Performance Analysis}
Table~\ref{tab:final_results} presents our final results on the EXIST 2025 test set, evaluated using the Information Contrast Measure (ICM-Hard)~\cite{amigo2022evaluating}, which provides a robust assessment of model performance under class imbalance and hierarchical label structures.

\begin{table}[htbp]
\centering
\caption{Final Results on EXIST 2025 Test Set on Hard Label Evaluation on ICM-Hard}
\label{tab:final_results}
\begin{tabular}{@{}lcccc@{}}
\toprule
\textbf{Subtask} & \textbf{English} & \textbf{Spanish} & \textbf{Overall}& \textbf{Overall Ranking} \\
\midrule
1.1 (Binary Sexism Detection) & 0.6231 & 0.7124 & 0.6774& 1 \\
1.2 (Source Intention Detection) & 0.3676 & 0.5937 & 0.4991 & 1 \\
1.3 (Sexism Categorization) & 0.5085 & 0.7650 & 0.6519 & 1 \\
\bottomrule
\end{tabular}
\end{table}

Our approach achieved first place across all three subtasks, showing the effectiveness of hierarchical LoRA adaptation with comprehensive module targeting.
Spanish consistently outperforms English across all subtasks, with particularly pronounced differences in intention detection ($+0.23$ ICM-Hard) and sexism categorization ($+0.26$ ICM-Hard). This cross-lingual performance gap likely reflects both differences in training data distribution and linguistic characteristics of sexist expressions across languages, suggesting that sexist discourse may manifest through more identifiable patterns in Spanish social media text.

As expected, performance decreases with task complexity, from binary classification (0.6774) to the more nuanced intention detection task ($0.4991$). This progression aligns with the inherent difficulty of fine-grained semantic understanding required for distinguishing between direct sexism, reported experiences, and judgmental commentary. Interestingly, the multilabel categorization task achieves intermediate performance ($0.6519$), suggesting that our hierarchical approach effectively leverages parent-level predictions to guide more complex downstream classifications.

The strong performance on hierarchically dependent tasks validates our design choice of conditional LoRA routing. Despite the challenging nature of intention detection -- which requires understanding pragmatic context and author stance -- our model maintains competitive performance by conditioning adapter selection on binary sexism predictions. This shows that explicitly modeling label dependencies through hierarchical specialization provides tangible benefits for complex, structured classification scenarios in social media discourse analysis.

\subsection{Efficiency Analysis and Ablation Study}

Our LoRA-based approach achieves substantial computational efficiency compared to full fine-tuning while maintaining competitive performance. Table~\ref{tab:efficiency} demonstrates that our method reduces trainable parameters by 98.33\% (from 8.03B to 134M), enabling training on consumer-grade GPUs with only 12GB memory compared to the 32GB required for full fine-tuning. This efficiency translates to 4x faster training times and 64x smaller storage footprint per task-specific adapter, making the deployment practical and cost-effective.

\begin{table}[t]
\centering
\caption{Computational Efficiency Comparison}
\label{tab:efficiency}
\begin{tabular}{@{}lcc@{}}
\toprule
\textbf{Metric} & \textbf{Full Fine-tuning} & \textbf{LoRA (Ours)} \\
\midrule
Trainable Parameters & 8.03B (100\%) & 134M (1.67\%) \\
GPU Memory (Training) & 32GB & 12GB \\
Training Time (per task) & 24 hours & 6 hours \\
Storage per Adapter & 32GB & 512MB \\
\bottomrule
\end{tabular}
\end{table}

To validate our hyperparameter selection, we conducted ablation studies examining the relationship between LoRA rank and model performance. Table~\ref{tab:lora_ablation} shows results on Subtask 1.1, revealing that rank 16 achieves optimal efficiency-performance trade-offs. While higher ranks (32, 64) yield marginal F1 improvements of less than 0.3\%, they require 2-4x more parameters and proportionally increased memory and training time. Our chosen configuration thus maximizes accessibility for researchers with limited computational resources while achieving near-optimal performance across all subtasks.

\begin{table}[t]
\centering
\caption{LoRA Rank Ablation Study on Subtask 1.1 with performance evaluated on validation set.}
\label{tab:lora_ablation}
\begin{tabular}{@{}cccc@{}}
\toprule
\textbf{Rank} & \textbf{Alpha} & \textbf{F1 Score} & \textbf{Parameters} \\
\midrule
8 & 8 & 0.851 & 67M \\
16 & 16 & \textbf{0.868} & 134M \\
32 & 32 & 0.869 (+0.1\%) & 268M \\
64 & 64 & 0.871 (+0.3\%) & 536M \\
\bottomrule
\end{tabular}
\end{table}

These efficiency gains are particularly crucial for the hierarchical multi-task nature of the shared task, where separate adapters for each subtask would traditionally require 3x the storage and memory. Our approach enables deployment of all three task-specific models within the memory constraints of a single GPU.

\section{Conclusion}
This paper presents a simple yet highly effective approach to text-based sexism detection that achieved first-place performance in EXIST 2025 Task 1 across English and Spanish languages on hard label evaluations. Our methodology demonstrates that straightforward Low-Rank Adaptation (LoRA) fine-tuning of Llama 3.1 8B, combined with unified multilingual training, outperforms more complex approaches while maintaining computational efficiency.

Our key finding is that joint bilingual training consistently surpasses separate language-specific models, achieving 1.7-2.4\% improvements across all subtasks. The bidirectional knowledge transfer between English and Spanish shows that shared semantic representations of sexist patterns can transcend language boundaries while preserving language-specific nuances. This finding has broader implications for multilingual classification tasks beyond sexism detection, extending previous work on cross-dataset transfer learning~\citep{Yuan2024,Yuan2023} to the cross-lingual domain.

While our approach achieves strong performance through straightforward supervised learning, incorporating demographic information into the fine-tuning process presents an opportunity for improvement. The EXIST 2025 dataset includes rich annotator demographic information, including gender, age, education level, ethnicity, and country of residence. 
Rather than discarding this valuable information in favor of gold labels, future work could explore encoding these persona characteristics directly into the LoRA adaptation process. This could involve persona-specific adapters~\cite{magnosao2025effects}, demographic-aware attention mechanisms, or multi-task learning approaches that jointly optimize for sexism detection while modeling annotator perspectives. Such persona-aware fine-tuning could capture the subjective nature of sexism perception across different demographic groups, leading to more culturally-sensitive detection systems.

Furthermore, our hierarchical LoRA approach opens avenues for integration with broader harmful content moderation frameworks~\citep{Schneider2023} and opinion market models~\citep{Calderon2024a} that could provide real-time intervention capabilities. Recent advances in early engagement prediction~\citep{Tian2025ICMamba} suggest that combining our sexism detection approach with temporal engagement forecasting could enable proactive identification of potentially viral sexist content within the critical first minutes of posting. Additionally, incorporating causal modeling approaches~\citep{Tian2025Causal} could help distinguish between organic engagement and coordinated amplification of sexist content, providing deeper insights into the true influence mechanisms underlying gender-based harassment campaigns. Future work could explore how our sexism detection system might be combined with positive intervention strategies to not only identify sexist content but also guide counter-narrative generation or targeted educational interventions, contributing to more comprehensive approaches for fostering equitable online discourse within the broader ecosystem of harmful content mitigation.

\bibliography{ref}

\end{document}